\documentclass[lettersize,journal]{IEEEtran}
\usepackage{amsmath,amsfonts}
\usepackage{algorithmic}
\usepackage{algorithm}
\usepackage{array}
\usepackage[caption=false,font=normalsize,labelfont=sf,textfont=sf]{subfig}
\usepackage{textcomp}
\usepackage{stfloats}
\usepackage{url}
\usepackage{verbatim}
\usepackage{graphicx}
\usepackage{cite}
\usepackage{booktabs}
\usepackage{bbm}
\usepackage{bm}
\usepackage{amssymb}
\usepackage{cite}
\usepackage{multirow}
\usepackage[accsupp]{axessibility}
\usepackage[pagebackref,breaklinks,colorlinks]{hyperref}
\usepackage{orcidlink}

\begin{document}

\title{MOWA: Multiple-in-One Image Warping Model} 

\author{Kang Liao, Zongsheng Yue, Zhonghua Wu, and Chen Change Loy,~\IEEEmembership{Senior Member,~IEEE}
        % <-this % stops a space
\thanks{This work was supported by the RIE2020 Industry Alignment Fund Industry Collaboration Projects (IAF-ICP) Funding Initiative, as well as cash and in-kind contribution from the industry partner(s). (\textit{Corresponding author: Chen Change Loy.})}
\thanks{Kang Liao, Zongsheng Yue, and Chen Change Loy are with the S-Lab, Nanyang Technological University (NTU), Singapore  (e-mail: kang.liao@ntu.edu.sg, zsyzam@gmail.com, and ccloy@ntu.edu.sg).}
\thanks{Zhonghua Wu is with the SenseTime Research (e-mail: wuzhonghua@sensetime.com).}
}

\markboth{IEEE TRANSACTIONS ON PATTERN ANALYSIS AND MACHINE INTELLIGENCE}
{Shell \MakeLowercase{\textit{et al.}}: A Sample Article Using IEEEtran.cls for IEEE Journals}

\maketitle

\begin{figure*}
    %\vspace{-6mm}
    \centering
    \includegraphics[width=.94\linewidth]{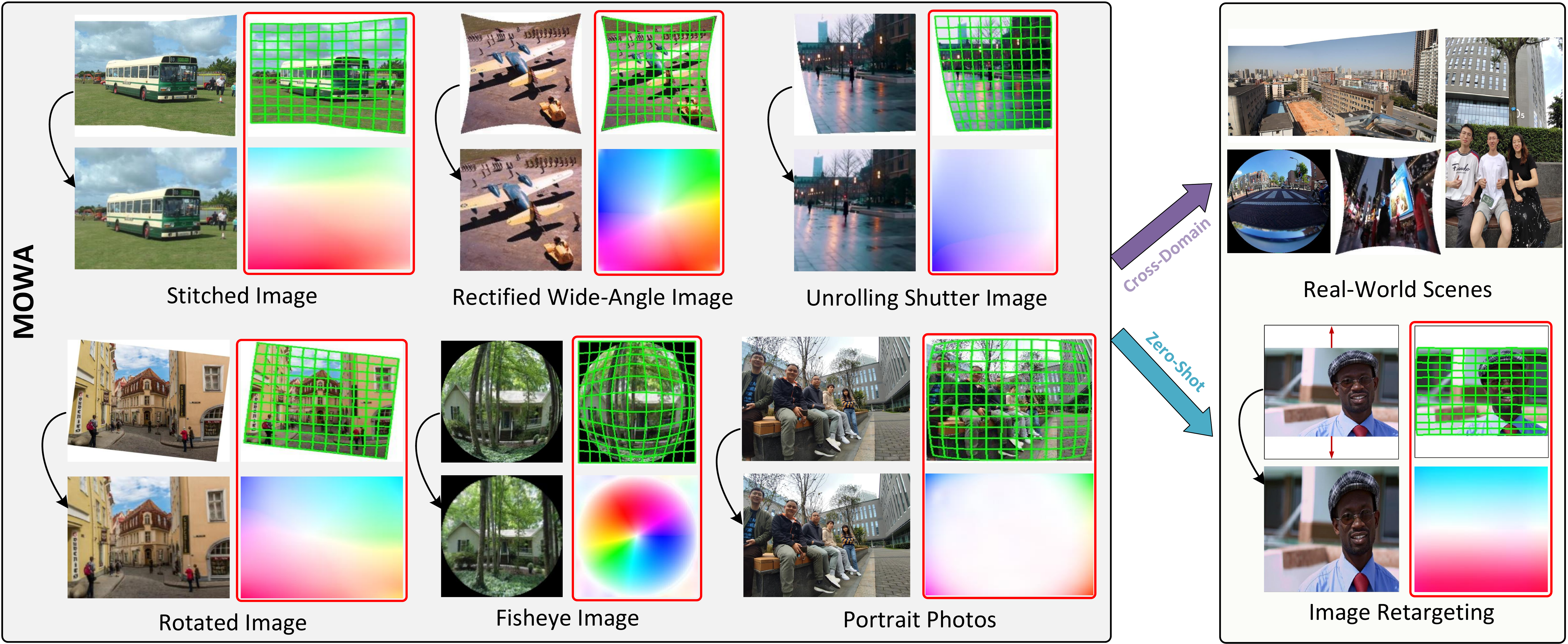}
    %\vspace{-5mm}
    \caption{MOWA is devised to address a variety of practical image warping tasks within a single framework, particularly in computational photography, where six distinct types of distortions are considered in this study. It also demonstrates an ability to generalize to novel scenarios, as evidenced in both cross-domain (unfamiliar domains) and zero-shot (unseen tasks) evaluations. The approach notably identifies and uses region-level and pixel-level fields, highlighted by red boxes, to accurately warp input images.
    }
    \label{fig:tesear}
    %\vspace{-10mm}
\end{figure*}

\begin{abstract}
While recent image warping approaches achieved remarkable success on existing benchmarks, they still require training separate models for each specific task and cannot generalize well to different camera models or customized manipulations. To address diverse types of warping in practice, we propose a \underline{M}ultiple-in-\underline{O}ne image \underline{WA}rping model (named \textbf{MOWA}) in this work. Specifically, we mitigate the difficulty of multi-task learning by disentangling the motion estimation at both the region level and pixel level. 
To further enable dynamic task-aware image warping, we introduce a lightweight point-based classifier that predicts the task type, serving as prompts to modulate the feature maps for more accurate estimation.
To our knowledge, this is the first work that solves multiple practical warping tasks in one single model. Extensive experiments demonstrate that our MOWA, which is trained on six tasks for multiple-in-one single image warping, outperforms state-of-the-art task-specific models across most tasks. Moreover, MOWA also exhibits promising potential to generalize into unseen scenes, as evidenced by cross-domain and zero-shot evaluations. The code and more visual results can be found on the project page: \url{https://kangliao929.github.io/projects/mowa/}.
\end{abstract}


\section{Introduction}
\IEEEPARstart{I}{mage} warping is essential in the field of computational imaging and computer vision, serving as the foundation for numerous applications, including image rectification~\cite{sawhney1999true, hartley2007parameter, feng2023simfir, wang2023model}, image rectangling~\cite{he2013rectangling, nie2022deep, li2015geodesic, zhang2020content}, camera calibration~\cite{kannala2006generic, jp, herrera2012joint, ramalingam2016unifying, liao2023deep}, and 3D reconstructions~\cite{hartley2003multiple, liu2017single, hane2016dense, kumar2019superpixel}, etc. Enabling the manipulation of image data through processes such as scaling, rotation, and shearing allows for the seamless integration of diverse visual elements and the correction of optical imperfections. Moreover, image warping is indispensable in developing augmented reality (AR) and virtual reality (VR) applications~\cite{ping2019comparison, thanyadit2019investigating, genay2021being}, where it helps create immersive and realistic environments by accurately mapping textures and images onto 3D models.

Considering different inputs derived from different camera models or manipulation spaces, recent works integrate specific prior knowledge into their models to address the corresponding image warping tasks~\cite{nie2022deep, he2013rectangling, recrecnet, wang2023model, feng2022geometric, feng2021docscanner, he2013content, tan2021practical, zhu2022semi}. While these single-task approaches achieve significant progress, we found they suffer from two main limitations: (i) the lack of generalization and flexibility, which restricts their real-world applications since users are required to manually identify each input type and apply the appropriate single-task model. This process is time-consuming and challenging for non-professional users to judge; (ii) the substantial storage requirements for multiple task-specific models, which are impractical for some resource-limited platforms. Thus, it is crucial to develop a holistic framework capable of efficiently warping images from various camera models or manipulation spaces. Furthermore, many image warping tasks typically involve shared processes, such as motion estimation and content-aware perception. This indicates the possibility of developing a unified framework that incorporates these common image techniques.

In this work, we propose a \underline{M}ultiple-in-\underline{O}ne image \underline{WA}rping model (named \textbf{MOWA}) to address various tasks in practice, as shown in Fig.~\ref{fig:tesear}. Specifically, we consider six representative types in the field of computational photography, namely stitched images, rectified wide-angle images, unrolling shutter images, rotated images, fisheye images, and portrait photos, covering the mainstream practical image warping tasks. 

Given the fact that learning different structures of motion is non-trivial in one model and motion representations differ significantly across various tasks, we propose to disentangle the motion estimation at both the region level and pixel level. In this hierarchical architecture, we first estimate the control points of the thin-plate spline (TPS) model~\cite{tps} with increasing refined numbers, in which the feature maps are progressively warped and rectified. Such a representation excels in approximating complex motions at region level \footnote{The region specifically denotes each grid formed by TPS control points rather than the segmented parts in some vision tasks.} and enables high flexibility to various motion structures. Subsequently, the warped feature maps are fed into the decoder to predict the residual pixel-level displacement, which further improves the warping results for each task, especially in the image boundaries and details.

To enable MOWA to explicitly discriminate diverse input types, a lightweight point-based classifier is devised. Adding an extra classification network based on the image features is a straightforward solution but brings high computation and storage costs. Noticing the motion structures in different tasks possess their specific distribution, we leverage the middle product of the image warping framework, \textit{i.e.}, region-level control points, to directly learn the task type. It achieves comparable performance while allowing significant parameter reduction compared to the image-based classifier since only a few 2D points are needed. Then, the task label predicted by this point-based classifier is used to modulate the feature maps in the decoder, dynamically boosting task-aware image warpings using a prompt learning module. Prompts are a set of learnable parameters that encapsulate essential discriminative information about different types of input, which empower a single model to efficiently traverse and harness its vast parameter space to accommodate various warping requirements.

In the experiments, we trained MOWA on six typical tasks for multiple-in-one single image warping. Experimental results demonstrated that it outperforms state-of-the-art (SotA) task-specific models in most tasks, even with comparable network parameters. In addition, MOWA allows the ability to generalize to unseen scenes, as evidenced by cross-domain evaluation (unfamiliar domains) and zero-shot evaluation (unseen tasks), indicating its robustness and adaptability across various scenarios. Our contributions can be summarized as follows:

\begin{itemize}
    \item We propose MOWA, which is the first practical multiple-in-one image warping framework. This proposed model, despite with an affordable model size, still evidently outperforms most SotA methods.
    
    \item We propose to mitigate the difficulty of multi-task learning by decoupling the motion estimation at both the region level and pixel level. Moreover, a prompt learning module, guided by a lightweight point-based classifier, is designed to facilitate task-aware image warpings.

    \item We show that through multi-task learning, our framework develops a robust generalized warping strategy that gains improved performance across various tasks and even generalizes to unseen tasks.
    
 \end{itemize}

The remainder of the paper is organized as follows: Section~\ref{sec:related} reviews the related works. We then present the proposed MOWA in Section~\ref{sec:method}. The experiments are provided in Section~\ref{sec:experiments}. Section~\ref{sec:conclusion} concludes this paper.

\section{Related Work}
\label{sec:related}
Image warping is the process of manipulating an image to change its shape or alignment. This transformation is achieved by applying a spatial mapping function to the coordinates of the original image, resulting in a new image with altered geometry. In computational photography, image warping is a key technique for enhancing and manipulating images beyond traditional photography limits. This technique enables the creation of panoramic images~\cite{brown2007automatic, lin2015adaptive, zhang2014parallax}, the correction of lens distortions~\cite{liao2019dr, bogdan2018deepcalib, kannala2006generic}, and the synthesis of novel views~\cite{zhou2016view, liu2018geometry, daribo2010depth}, etc. In the past few decades, warping techniques have significantly contributed to the development of advanced imaging applications beyond those mentioned above, offering greater flexibility and creativity.
For example, the image boundaries can be twisted by different manipulations, leading to visually unpleasant layouts and negative effects on downstream vision tasks. Nevertheless, in practical scenarios, most users favor rectangular boundaries due to their compatibility with standard display formats, facilitating ease of sharing, printing, and publication~\cite{he2013rectangling, he2013content}. Therefore, researchers have developed diverse image rectangling methods to warp the image boundaries to be straight~\cite{he2013rectangling, nie2022deep, recrecnet, li2015geodesic, zhang2020content, zhou2023rectangular}. Most of them follow the principle of content-aware image warping to avoid the large distortion on the original distribution when rectangling the image. Besides, different motion representations are also exploited, such as the mesh~\cite{he2013rectangling, nie2022deep} and control points~\cite{recrecnet}, to formulate the warping process.

Excluding the customized manipulations, some special camera models can introduce geometric distortion onto the captured images, $e.g.$, radial distortion, rolling shutter distortion, and perspective distortion. The images' semantic features significantly disobey the real-world rules due to those distortions. To address this issue, there is an exploration of distortion correction approaches~\cite{liao2019dr, bogdan2018deepcalib, yang2021progressively, feng2023simfir, yan2023deep, liu2020deep, rengarajan2017unrolling, tan2021practical, shih2019distortion, zhu2022semi, wang2023model} aimed at warping the distorted input to a geometrically reasonable one. Particularly, regression-based methods~\cite{bogdan2018deepcalib, rengarajan2017unrolling} learn the camera and distortion parameters from the input image and correct the distortion by simulating the imaging process of a predefined camera model. In contrast, reconstruction-based methods~\cite{liao2019dr, yang2021progressively, feng2023simfir, yan2023deep, tan2021practical, liao2023dafir} directly learn the pixel-wise displacement between the distorted image and its ground truth, facilitating the model-free correction and enabling the end-to-end training.

The above works achieve remarkable progress on various tasks, of which well-designed network architectures and tailored motion representations are studied. However, they need to train an individual model for each specific warping type and require prior knowledge of the camera model or customized manipulations. In this work, we propose a multiple-in-one framework to involve these typical and practical image warping tasks. We address the challenge of learning different motion structures within a single model by employing a coarse-to-fine approach, progressively adding more TPS points to accurately fit the expected geometric distribution. To compensate additional degrees of freedom for TPS, our method further learns the residual flow based on the warped feature map, allowing for tuning of image boundaries and details in the final results.

\begin{figure*}[t]
    \centering
    \includegraphics[width=.9\linewidth]{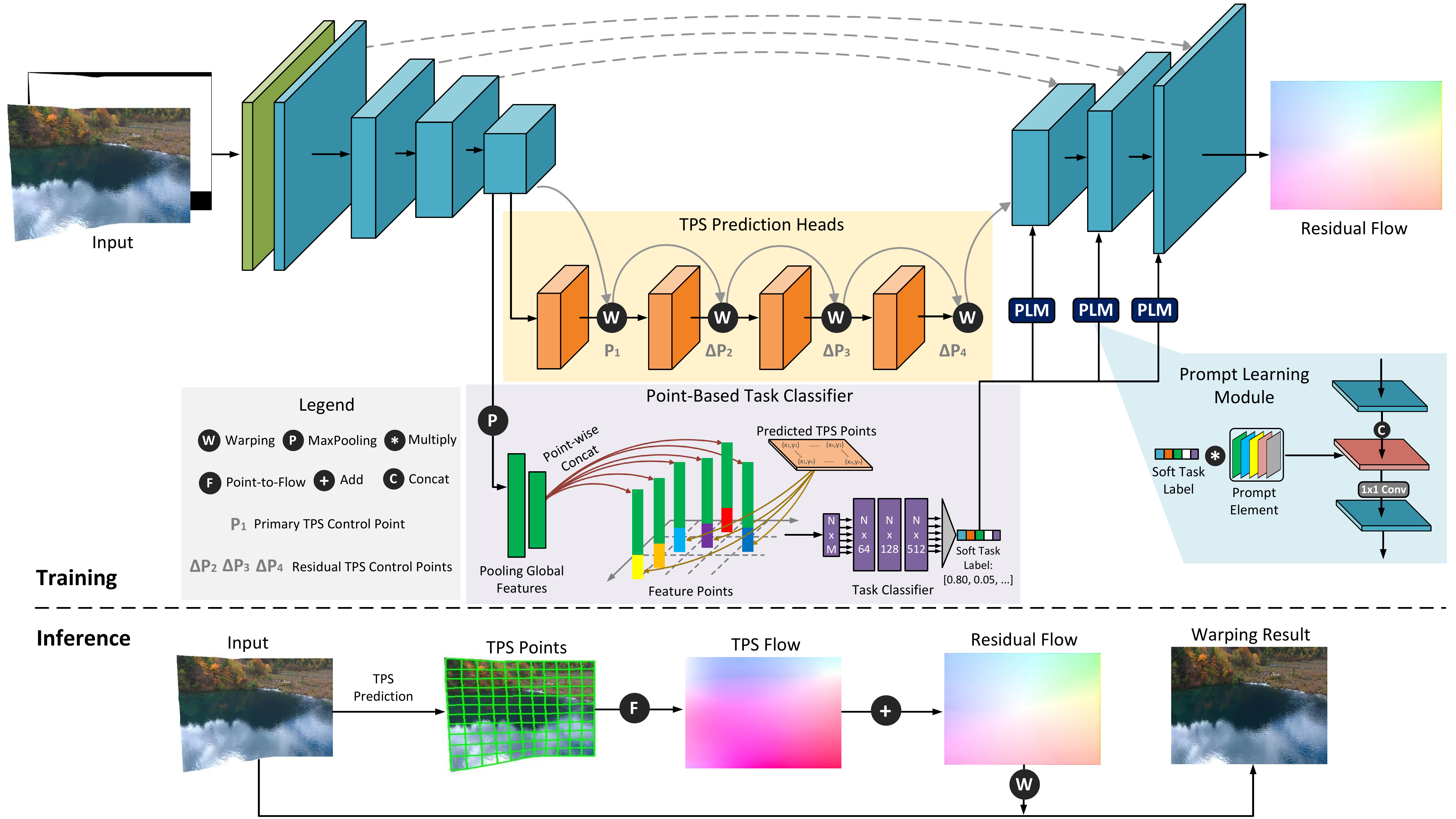}
    \caption{Overview of the proposed multiple-in-one image warping model (MOWA). It begins by taking an image and a mask as input to estimate the TPS control points with progressively refined precision. During such a region-level motion estimation, feature maps are incrementally warped and rectified. These warped features are then passed to the decoder to predict residual pixel-level motion. To ensure task awareness and expandability, a lightweight point-based classifier and a prompt learning module are designed. During inference, MOWA supports image warping for any resolution by scaling the predicted TPS control points and residual flow.
    }
    \label{fig:framework}
    %\vspace{-0.4cm}
\end{figure*}

\section{Multiple-in-One Warping Model}
\label{sec:method}

\subsection{Problem Definition}
In this study, we consider six representative and practical image types in the field of computational photography, including stitched images, rectified wide-angle images, unrolling shutter images, rotated images, fisheye images, and portrait photos, covering the mainstream practical image warping tasks. These types are further classified into two groups. The first four types (stitched images, rectified wide-angle images, unrolling shutter images, and rotated images) struggle with irregular boundaries as the original images are manipulated by some customized operations, such as image stitching, distortion correction, and rotation. Therefore, image rectangling is proposed to reshape these irregular boundaries while keeping the distribution of content unchanged. The last two types (fisheye images and portrait photos) show inherently geometric distortions imaged by special camera models, such as radial distortion in fisheye images and multiple distortions (both radial distortion and perspective distortion) in portrait photos. Correcting these distortions is crucial to scene understanding and aesthetic appreciation. 

Figure~\ref{fig:framework} shows the overall framework of the proposed MOWA. It takes the image and mask as input and estimates the TPS control points with increasingly refined numbers. In this region-level motion estimation, the feature maps are progressively warped and rectified. Subsequently, the warped features are fed into the decoder to predict a residual pixel-level motion. To enable task-aware and expandable capabilities, a lightweight point-based classifier and prompt learning module are designed. We elaborate on the details of each module in MOWA as follows. 

\begin{figure*}[t]
    \centering
    \includegraphics[width=.95\linewidth]{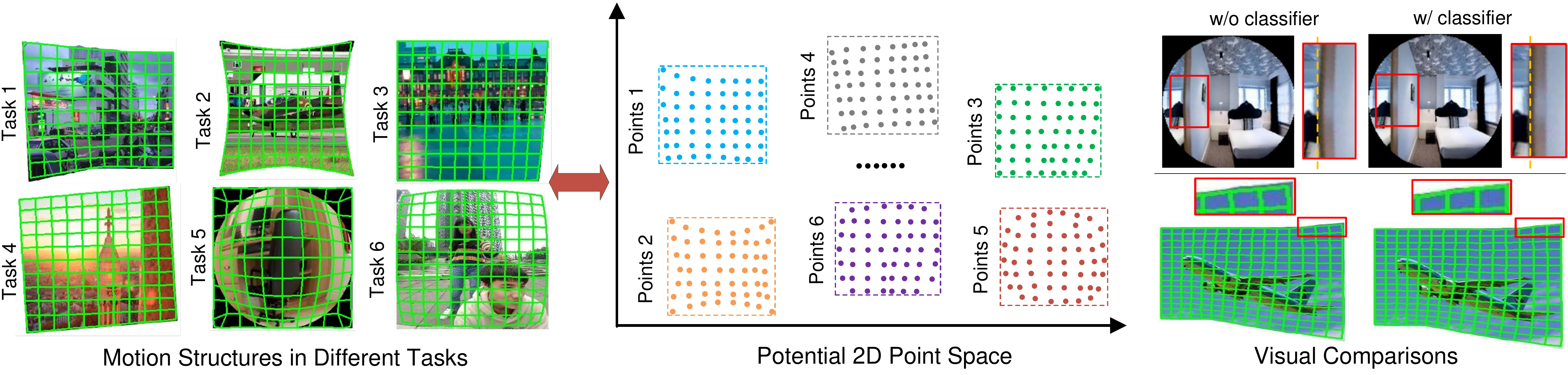}
    \caption{Motion structures in different tasks possess their specific distribution, which potentially exists in a 2D point space. Discriminating these motion structures as a classification task can also help the image warping performance as exhibited in visual comparisons.}
    \label{fig:point_structure}
    %\vspace{-0.4cm}
\end{figure*}

\subsection{Motion Estimation Module}
Learning multiple warping types in one model is challenging since the network needs to balance different complexities of the multiple motion types in motion estimation. Furthermore, the model's scalability would be restricted if the motion representation is hand-crafted for specific tasks. Hence, we propose a flexible and hierarchical architecture for general image warping in MOWA. As shown in Fig.~\ref{fig:framework}, the motion estimation is disentangled at both the region level, where the number of TPS control points progressively increases, and the pixel level, where a residual map is predicted to further compensate the estimated TPS flow.

\noindent\textbf{Region-Level Motion Estimation.} The TPS transformation~\cite{tps} stands out for its remarkable ability to model complex motions~\cite{detlefsen2018deep, rocco2017convolutional, li2020structured}. It is adept at performing image warping based on two sets of region-level control points, namely $\bm{Q} = [\bm{q}_1, \bm{q}_2, \cdots, \bm{q}_N]^T \in \mathbb{R}^{N\times 2} $ for the source image and $\bm{Q}^{'} = [\bm{q}'_1, \bm{q}'_2, \cdots, \bm{q}'_N]  \in \mathbb{R}^{N\times 2}$ for the target image. To minimize the distortion of the source and target images, an energy term is introduced to penalize the Euclidean distance between the transformed source points $\mathcal{T}(\bm{q}_{i})$ and the target points $\bm{q}'_{i}$, \textit{i.e.}, $\sum_{i=1}^{N} \lVert \mathcal{T}({\bm{q}_{i}})- {\bm{q}'_{i}}\rVert_{2}^{2}$.
This penalty results in a spatial deformation function parameterized by the control points, effectively capturing the intricate deformations across the image and maintaining the overall structural integrity. Specifically, the derived spatial deformation function can be expressed as follows: 

\begin{equation}
\label{eq_tps_cp}
  \mathcal{T}(\bm{q})=\bm{A}\begin{bmatrix}
    \bm{q}\\ 
    1   
    \end{bmatrix}+\sum_{i=1}^{N} U\left({\left\lVert \bm{q}'_{i}-\bm{q}\right\rVert}_2\right) \bm{w}_{i} ,
\end{equation}
where $\bm{q}$ represents a point located in the source image. $\bm{A} \in \mathbb{R}^{2\times 3}$ and $\bm{w}_{i} \in \mathbb{R}^{2}$ are the transformation parameters, $U(\cdot)$ is a radial basis function to quantify the influence of the control point,  more details can be found in literature~\cite{tps}. 
Notably, this deformation function plays a key role in determining the deformation induced by each control point, thereby shaping the overall transformation.

Motion estimation acts as the fundamental stage in image warping, presenting particular challenges in the context of multi-in-one task learning. To enhance the capability of our model in motion estimation, we design a progressive motion estimation module. More specifically, this module cascades a sequence of TPS transformation heads that gradually increase the number of control points. The control points predicted by the preceding head are upsampled and integrated into the prediction of the next head. Subsequently, these control points are arranged to generate a mesh. Then we adopt the TPS transformation to warp this mesh, aiming to align it with the regular mesh defined on the ground truth image. In the implementation, considering the cascade of fully connected layers introduces significant computation and storage costs, we use one or two \textit{convolution layers} to predict the control points after each TPS transformation head. The pipeline of cascaded TPS transformation heads can be expressed by:

\begin{equation}
\label{eq_tps_heads}
\bm{q}^{(t)} = h^{(t)}[\mathcal{R}(\mathcal{F}^{(t-1)}, \bm{q}^{(t-1)})] + \text{UP}[\bm{q}^{(t-1)}],
\end{equation}
where $h^{(t)}$ is the $t$-th TPS transformation head, $\mathcal{F}^{(t-1)}$ and $\bm{q}^{(t-1)}$ are the feature map and control points of the $(t-1)$-th head, respectively. $\mathcal{R}(\cdot,\cdot)$ represents the warping operation for feature maps given control points, and $\text{UP}[\cdot]$ is a customized upsampling layer for control points.

\noindent\textbf{Pixel-Level Motion Estimation.} While TPS transformation is flexible and adaptable to various tasks, it is limited in its ability to describe detailed motions due to its restricted degree of freedom. To alleviate this limitation, we further complement the region-level motion representation with a pixel-level residual flow. Specifically, we first rectify the feature map $\mathcal{F}^{(T)}$ using the corresponding control points $\bm{q}^{(T)}$ in the last transformation head, and then feed the rectified feature map into a decoder network $\mathcal{D}[\cdot]$ to predict the desirable residual flow. Like common U-Net architectures, the shallow features in the encoder are transited into the decoder using skip-connection. To eliminate the blur effect by multiple warpings (interpolation operation involved), we densify the TPS control points to pixel level and couple it with the residual flow to directly warp the input image $\mathcal{I}$. The final warping result $\mathcal{I}'$ can be obtained by:
\begin{equation}
\label{eq_warp}
\mathcal{I}' = \mathcal{W}\left(\mathcal{D}[\mathcal{R}(\mathcal{F}^{(T)}, \bm{q}^{(T)})] + \text{DE}[\bm{q}^{(T)}], \mathcal{I}\right),
\end{equation}
where $\mathcal{W}(\cdot, \cdot)$ denotes the warping operation given the flow map and input image, $\text{DE}[\cdot]$ densifies the sparse control points to a dense flow map, which can be regarded as a special case of TPS upsampling layer $\text{UP}[\cdot]$. Unlike previous works tailored for specific warping tasks, our method unifies motion representation across various tasks at both the region and pixel levels. The experimental results are demonstrated in Section~\ref{sec:experiment_ablation}.
 
\subsection{Point-based Task Classifier}
When learning various image warping tasks simultaneously, a task classifier is crucial for efficiently routing inputs to their respective task-specific components, optimizing resource use, and enhancing model performance. It is straightforward to design a task classifier in terms of the input image. However, such a design brings an unavoidable issue of high computation complexity due to the redundant image features. Instead, we propose a lightweight task classifier based on the TPS points predicted by the motion estimation module. Our motivation stems from the fact that the motion structures in different tasks possess their specific distribution, which potentially exists in a point space as shown in Fig.~\ref{fig:point_structure}. To this end, we design a PointNet-like network~\cite{qi2017pointnet, qi2017pointnetpp} to predict the task type. Specifically, as shown in Fig.~\ref{fig:framework}, it takes the local coordinates of motion with the global image features (after maxpooling) by point-wise concatenation along the last dimension as input and outputs the soft task label $\bm{\Phi}$. We can formulate this point-based task classifier as follows:
\begin{align}
\mathcal{F}_g = f
\left[\text{MaxPool}(\mathcal{F})\right], \ \ \ \ \ \ \ \ \ \ \ \ \ \\
\bm{\Phi} = \text{Softmax}\left(f'[\bm{q}^{(T)} \oplus \text{R}(\mathcal{F}_g, [1, H \times W])]\right),
\end{align}
where $f$ and $f'$ are the fully connected layers to decrease the dimensions of features and learn the abstract concepts. $\text{R}$ denotes replicating $\mathcal{F}_g$ to the same shape of the predicted motion coordinates $\bm{q}^{(T)}$ and $\oplus$ is the point-wise concatenation. Experiments demonstrate our point-based task classifier achieves comparable results while having less than $\times 50$ parameters compared with the image-based classifier.

In addition to the task classification function, our point-based task classifier can further improve image warping performance. This improvement is due to the high-level guidance provided by the task classifier to the motion estimation module through gradient back-propagation. Figure~\ref{fig:point_structure} (right) depicts a typical example, in which the fisheye rectification result shows a less distorted shape, and the predicted control points of the stitched image are more tightly aligned to the image boundary. Compared to the vanilla baseline, the proposed point-based task classifier achieves an average improvement of $+0.35$dB in PSNR metrics across various image warping tasks. More quantitative results are presented in Section~\ref{sec:experiment_ablation}.

\subsection{Prompt Learning Module}
Once the inputs are classified by the proposed point-based network, we leverage the predicted task label to modulate the feature maps in the network. In particular, a prompt learning block is inserted into each layer in the decoder as a plug-and-play module. Prompt learning aims to tackle the challenge of generalizing in various image warping tasks by aiding the network in comprehending the specific task at hand. The prompts serve as a flexible and lightweight component to encode motion context across multiple scales within the image warping network.

Assuming the task number is $N$, we introduce a set of learnable parameters as our prompts, namely $\{\bm{P}_i\}_{i=1}^N$. By denoting the predicted task label by the task classifier as $\bm{\Phi}$, we modulate the feature maps $\mathcal{F}$ in the decoder network by the prompts as follows:
\begin{equation}\label{eq_prompt}
  \mathcal{F}^{m} = \text{Conv}_{1\times 1}\big(\mathcal{F} \oplus \sum_{i=1}^N{\Phi_i \bm{P}_i}\big),
\end{equation}
where $\oplus$ represents the concat operation, $\text{Conv}_{1\times 1}$ is a convolution layer with $1\times 1$ kernel size aiming to reduce the channel dimension of concatenated features. 

By integrating the learnable prompts with the features of the warping model, we can significantly enrich the representations with task-specific knowledge. Unlike pre-defined and fixed prompts, our adaptive approach enables the network to dynamically influence its behavior, resulting in more efficient and precise image warping. This adaptive process not only enhances the flexibility of the model but also improves its ability to generalize across different tasks and datasets. More analysis on the multi-task learning and effectiveness of the proposed prompt learning are demonstrated in Section~\ref{sec:experiment_multitask}.

\subsection{Training Loss}
After predicting the TPS control points and the residual flow, the warped image can be obtained by Eq.~\eqref{eq_warp}. Following previous works~\cite{nie2022deep, recrecnet}, we first exploit three losses to train our multiple-in-one image warping framework, \textit{e.g.}, image reconstruction loss $\mathcal{L}_{Rec}$, perceptual loss $\mathcal{L}_{Per}$, and inter-grid loss $\mathcal{L}_{Grid}$. The reconstruction loss and perceptual loss supervise the warped image at the pixel level and feature level, respectively. The inter-grid loss constrains the edges of two consecutive deformed grids $\{\vec{e}_{t1}, \vec{e}_{t2}\}$ to be co-linear:
\begin{equation}\label{eq_loss_ig}
  \mathcal{L}_{Grid}= \frac{1}{M}\sum_{\{\vec{e}_{t1}, \vec{e}_{t2}\}\in m}(1-\frac{\langle \vec{e}_{t1},\vec{e}_{t2}\rangle}{\parallel \vec{e}_{t1}\parallel \cdot \parallel \vec{e}_{t2}\parallel }).
\end{equation}
Here, $M$ represents the number of tuples of two successive edges in a mesh $m$. When maximizing the above cosine representation, the corresponding two edges become collinear. Consequently, the loss reaches its minimum, ensuring the image content remains consistent.

Considering the ground truth of warping flow is available in the training dataset of portrait photos, we also add the reconstruction loss $\mathcal{L}_{Flow}$ on the predicted flow of the portrait correction task. Moreover, we provide middle-level supervision on the warped results from the TPS prediction heads with a set of exponentially growing weights. To train the point-based task classifier, the standard cross-entropy loss $\mathcal{L}_{Cls}$ is applied. Overall, the final loss can be expressed by:
\begin{equation}\label{eq_loss_all}
\mathcal{L} = \underbrace{\mathcal{L}_{Rec} + \mathcal{L}_{Per} + \mathcal{L}_{Grid} + \lambda_{Flow}\mathcal{L}_{Flow}}_{Image\  Warping} + \underbrace{\lambda_{Cls}\mathcal{L}_{Cls}}_{Task\  Classifier},
\end{equation}
where $\lambda_{Flow}$ and $\lambda_{Cls}$ are the hyperparameters to balance different losses, both of them are empirically set to $0.1$.

In summary, the proposed multiple-in-one image warping framework brings the following benefits.

\begin{itemize}
    \item Unlike previous task-specific image warping models, our method can recover various geometrically distortion images within a single network. It does not require prior knowledge of the camera models or manipulation spaces; it is also friendly to use and relies only on the observed input image to perform the customized image warping.
    \item Our method provides greater flexibility and cost-effectiveness in real-world scenarios, unlike previous methods that need a proportionally larger model size as the number of warping tasks increases.
    \item Thanks to multi-task learning, our method develops a generalized motion representation across various image warping tasks, demonstrating remarkable performance in cross-domain evaluations and unseen tasks.
\end{itemize}

\section{Experiments}
\label{sec:experiments}
To demonstrate the effectiveness of the proposed multiple-in-one image warping method, we evaluate its performance on six representative distorted types, including stitched images, rectified wide-angle images, unrolling shutter images, rotated images, fisheye images, and portrait photos, covering the mainstream practical image warping tasks.

\subsection{Experimental Settings}

\vspace{1mm}\noindent \textbf{Implementation Details}.
We train the proposed model using the AdamW optimizer with the momentum terms of $(0.9, 0.999)$ on 8 NVIDIA A100 GPUs. The learning rate starts with a linear warm-up in the first three epochs and then decays from $1e^{-4}$ to $1e^{-6}$ following a cosine schedule in the remaining epochs. The batch size is set as 64. The complete framework is trained with a fixed input size of $256\times256$. At the first 10 epochs, we solely train and supervise the TPS prediction heads with the point-based task classifier. Afterwards, all modules are trained collectively. During inference, the proposed method supports image warping for any resolution by scaling the predicted TPS control points and residual flow.

\vspace{1mm} \noindent \textbf{Network Configuration}. We design the image warping network based on the encoder-decoder architecture, enabling both region-level control point regression and pixel-level residual flow prediction. Specifically, the Transformer blocks with shifted windows~\cite{liu2021swin, wang2022uformer} are used in both the encoder and decoder except for the input projection layer and output projection layer. The basic dimension of channels is set to 32 and linearly increases along the layers in the encoder network, which is oppositely decreased to 2 in the decoder network. Moreover, the depths of each Transformer block are set to 2 and the head numbers of multi-head self-attention are $[1, 2, 4, 8, 16, 16, 8, 4, 2]$ along the whole layers. In TPS prediction heads, we adopt the convolution layers with different kernels to predict increasing numbers of control points, and the numbers are set to $10\times10$, $12\times12$, $14\times14$, and $16\times16$. The configuration details of these regression heads are listed in Table~\ref{tab:cp_layer}. Such a design enables significant parameter reduction compared with the fully connected layers. For the lightweight point-based classifier, three 1D convolutional layers with channel dimensions of $256, 256, 512$ are used to extract the features of input, and then three fully connected layers with unit numbers of $512, 256, 6$ are used to classify their task types.

\begin{table}[t]
\setlength{\tabcolsep}{3pt}
\begin{center}
\caption{The configurations of convolutional layers to regress different sizes of control points (the size of input feature maps is $16\times16$).}
\label{tab:cp_layer}
%\vspace{-3mm}
\footnotesize
\begin{tabular}{l|ccccc}
\hline
Configuration & $8\times8$ & $10\times10$ & $12\times12$ & $14\times14$ & $16\times16$\\
\hline
\hline
Kernel Size & $3\times3$ & \{$5\times5$, $3\times3$\} & $5\times5$ & $3\times3$  & $3\times3$\\
Stride & 2 & \{1, 1\} & 1 & 1 & 1\\
Padding & 1 & \{0, 0\} & 0 & 0 & 1\\
\hline
\end{tabular}
\end{center}
\footnotesize
%\vspace{-0.2cm}
\end{table}

\begin{figure*}[t]
    \centering
    \includegraphics[width=1\linewidth]{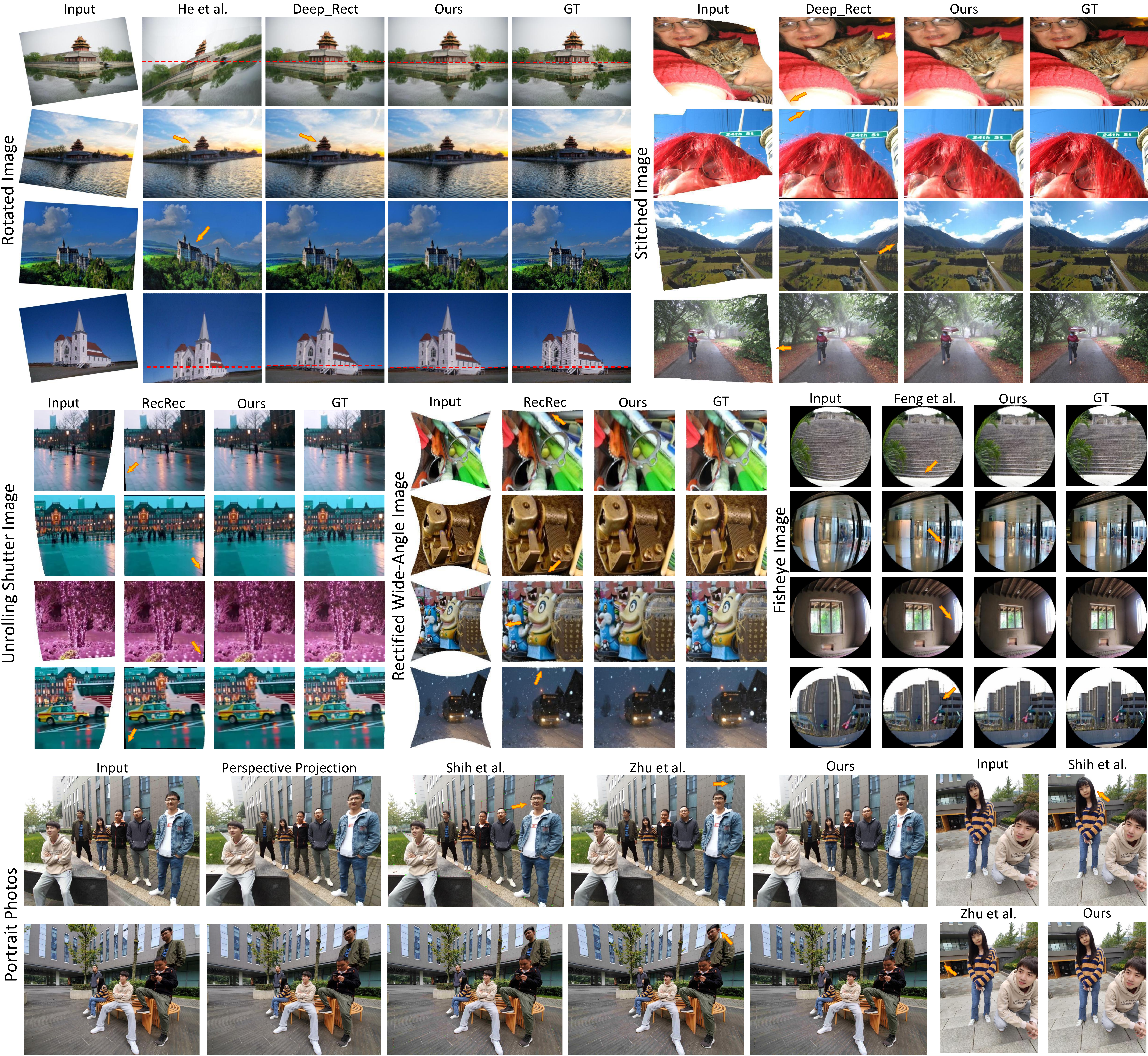}
    %\vspace{-0.5cm}
    \caption{Qualitative comparison of our multiple-in-one framework MOWA to the SotA image warping models. The red dotted lines mark the horizon. The arrows highlight the inferior warped parts such as the irregular boundaries and distorted semantics.}
    \label{fig:qualitative}
    %\vspace{-0.4cm}
\end{figure*}

\begin{table*}[h]
  \centering
  \caption{Quantitative evaluation of the proposed multiple-in-one framework to the SotA image warping models.}
  %\vspace{-0.3cm}
    \begin{tabular}{lcccccccc}
    \toprule
    \multicolumn{3}{c}{Warping Tasks} &       & \multicolumn{3}{c}{Metrics} \\
\cmidrule{1-3}\cmidrule{5-8}    \multicolumn{1}{c}{Input Type} &       & \multicolumn{1}{c}{Methods} &       & \multicolumn{1}{c}{PSNR $\uparrow$} & \multicolumn{1}{c}{SSIM $\uparrow$ } & \multicolumn{1}{c}{ShapeAcc $\uparrow$} & \multicolumn{1}{c}{Parameter}\\
\cmidrule{1-1}\cmidrule{3-3}\cmidrule{5-8}  

\multirow{3}*{Rotated Image} &       &   He et al.~\cite{he2013rectangling}    &       &  17.63     &   0.4880    &   - & - \\
 &       &   Deep\_Rect~\cite{nie2022deep}    &       &  19.89     &   0.5500    &   - & 52.14M \\
 &       &   Ours    &       &  21.01     &   0.5961    &   - &49.93M \\
\midrule
\multirow{3}*{Rectified Wide-Angle Image} &       &   He et al.~\cite{he2013rectangling}    &       &   15.36     &    0.4211    &   - & -\\
 &       &   RecRecNet~\cite{recrecnet}    &       &  18.68     &   0.5450    &   - & 62.70M\\
 &       &   Ours    &       &  18.69     &   0.5450    &   - & 49.93M\\
\midrule
\multirow{3}*{Stitched Image} &       &   He et al.~\cite{he2013rectangling}    &       &  14.70     &   0.3775    &   - &   -\\
 &       &   Deep\_Rect~\cite{nie2022deep}    &       &  21.28     &   0.7140    &   - &   52.14M\\
 &       &   Ours    &       &  20.72     &   0.6425    &   - &   49.93M\\
\midrule
\multirow{2}*{Unrolling Shutter Image} &       &   RecRecNet~\cite{recrecnet}    &       &  21.48     &   0.7602    &   - & 62.70M\\
 &       &   Ours    &       &  21.69     &   0.7795    &   - & 49.93M\\
\midrule
\multirow{3}*{Fisheye Image} &       &  PCN~\cite{yang2021progressively}    &       &  21.37     &   0.6925    &   - & 26.19M\\
 &       &   Feng et al.~\cite{feng2023simfir}    &       &  21.72     &   0.7167    &   - & 11.65M\\
 &       &   Ours    &       &  22.25     &   0.7488    &   - & 49.93M\\
\midrule
\multirow{4}*{Portrait Photos} &       &   Shih et al.~\cite{shih2019distortion}    &      &    -    &     -   &   97.253  & -\\
 &       &   Tan et al.~\cite{tan2021practical}    &       &     -   &       - &   97.490 & - \\
  &       &   Zhu et al.~\cite{zhu2022semi}    &       &     -   &       - &   97.491 & 8.79M\\
 &       &   Ours    &       &   -     &    -    &   97.477 & 49.93M\\
\bottomrule
    \end{tabular}
  \label{tab:quantitative}
%\small
%\vspace{-0.1cm}
\end{table*}

\vspace{1mm} \noindent \textbf{Datasets}.
We use the public benchmarks from recent SotA works, including the image rectangling datasets~\cite{nie2022deep, nie2023deep, recrecnet} and the distortion correction datasets~\cite{yang2021progressively, tan2021practical, zhu2022semi}. Since there is no available training dataset for unrolling shutter image rectangling, we use the rolling shutter correction dataset~\cite{yan2023deep} and follow the standard data construction process from previous methods~\cite{nie2022deep} to synthesize the paired data. This dataset will also be made public.
 
\vspace{1mm} \noindent \textbf{Metrics}. Following previous works, we select PSNR and SSIM as metrics to quantitatively measure the quality of the warped results. Please note that it is challenging to use the Average Endpoint Error (EPE) metric to evaluate the motion estimation performance in practical image warping tasks, because the accurate labels of motion are hard to obtain and unavailable in all the above test datasets. As a consequence, most previous methods have opted to learn the motion in an unsupervised manner and supervise the image warping model at the warped pixel level.

For the portrait correction task, the ShapeAcc metric is applied as suggested in Tan et al.~\cite{tan2021practical}. It is specially designed for the quality of face correction, which calculates the similarity between corrected portraits and the stereographic projection of its original input.

\subsection{Comparison Results}
We compare the proposed MOWA with recent SotA methods on each task, including Deep\_Rect~\cite{nie2022deep}, He et al.~\cite{he2013rectangling}, RecRecNet~\cite{recrecnet}, PCN~\cite{yang2021progressively}, Feng et al.~\cite{feng2023simfir}, Shih et al.~\cite{shih2019distortion}, Tan et al.~\cite{tan2021practical}, and Zhu et al.~\cite{zhu2022semi}.

\vspace{1mm}\noindent \textbf{Qualitative Comparison}.
As shown in Fig.~\ref{fig:qualitative}, we visualize the comparison results of different methods on the testing datasets. These qualitative results demonstrate that our multiple-in-one method can handle various tasks, scenes, and resolutions well, compared with the SotA methods specially designed for each task. For example, for the rotated images, our method can rearrange the input to a rectangle one while keeping the original geometric layout reasonable. On the contrary, distorted buildings can be observed in the results of previous works~\cite{he2013rectangling, nie2022deep}, in which the physical world rules such as the horizon are perturbed. For other rectangling tasks like the stitched image, unrolling shutter image, and rectified wide-angle image, our method shows a better visual appearance, especially in the image boundaries, allowing promising structural integrity among the comparison methods. In some challenging cases, such as the first and second rows in stitched images, the image boundaries are dramatically stretched, but our method can still warp the images to the expected structures. One important reason is that MOWA learns the generalized warping strategy from different tasks since it can extract some common knowledge from them.
In addition, our method mitigates the difficulty of motion estimation by disentangling it at both the region level and pixel level. Consequently, diverse structures of motions can be progressively approximated and the image details can be preserved. For the fisheye image and portrait photos, MOWA is capable of recovering the realistic distribution from the inputs with distorted objects or irregular boundaries, despite the radial distortion or perspective distortion. Please refer to more visual comparison images, interactive warping visualizations, and dynamic warping results on the project page: \url{https://kangliao929.github.io/projects/mowa/}.

\vspace{1mm}\noindent \textbf{Quantitative Comparison}.
We report the quantitative evaluation results in Table~\ref{tab:quantitative}. The proposed multiple-in-one image warping jointly learns six tasks and achieves promising performance compared with the single-task methods. For example, MOWA outperforms the SotA methods in rectified wide-angle images, unrolling shutter images, rotated images, and fisheye images, thanks to the elaborately designed hierarchical motion estimation architecture and task-aware prompt learning strategy. 
Moreover, MOWA achieves comparable image warping performance for stitched images and portrait photos without intolerable performance degradation when involving more tasks and data. The results suggest the generalizability and flexibility of MOWA, which are not achievable by previous methods~\cite{recrecnet, he2013rectangling, feng2023simfir, nie2022deep, tan2021practical} that tailor the specific knowledge into their models to address the single image warping task.

\noindent \textbf{Computation Complexity Comparison}.
In Table~\ref{tab:quantitative}, we also compare the computation complexity of the proposed method with previous methods that make their models available. The comparison suggests that our model size is reasonable and affordable as a multiple-in-one image warping framework. Even compared to the SotA models designed for the specific task~\cite{nie2022deep, recrecnet}, our MOWA has fewer parameters to achieve better or comparable warping performance. The underlying reason is that the shared knowledge across different tasks can relieve the burden of the parameter requirements of a multi-task model. Besides, the proposed motion estimation module discards the heavy fully connected layers and replaces them with convolutional layers. Then, the predicted region-level TPS points are further compensated with the pixel-level displacement from a compact convolutional decoder.

\begin{figure*}[t]
    \centering
    \includegraphics[width=1\linewidth]{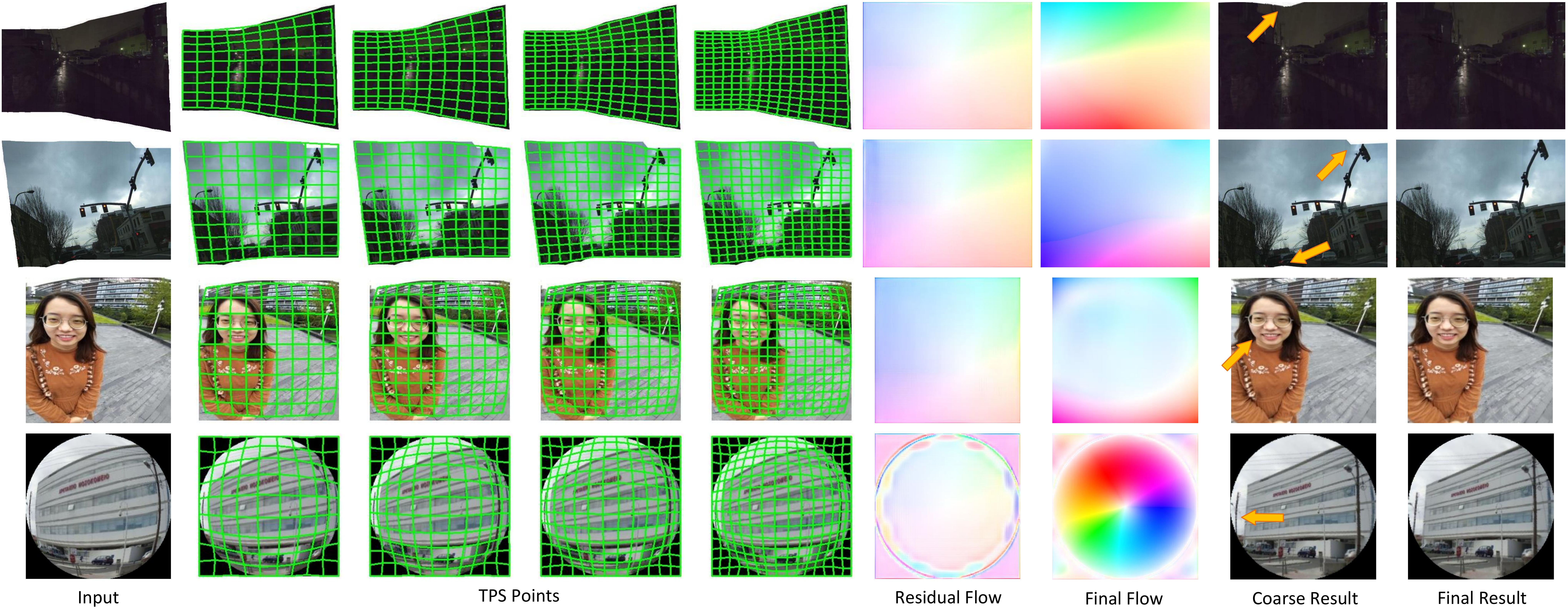}
     \vspace{-0.6cm}
    \caption{Ablation study on the proposed motion estimation module. The predicted TPS control points are shown with the size of $10\times10$, $12\times12$, $14\times14$, and $16\times16$, from left to right. The coarse results and final results are obtained by warping the input using the first control points and final flow (coupled with the last TPS points and residual flow), respectively.}
    \label{fig:motion_ablation}
    %\vspace{-0.2cm}
\end{figure*}

\subsection{Ablation Study}
\label{sec:experiment_ablation}
Considering the aim of a multiple-in-one framework is to achieve holistic performance across various tasks, we mainly compare the different variants of the framework in terms of the average warping metrics. Additionally, the same image quality metrics (PSNR and SSIM) are shared in the first five tasks, but the portrait correction task has its own metrics like ShapeAcc. Thus, the average PSNR and SSIM from the first five tasks are mainly reported in this part.

\begin{table}[t]
\setlength{\tabcolsep}{1.2pt}
\small
\begin{center}
\caption{Ablation study on the proposed motion estimation module. The baseline represents the predicted control points with a size of $12\times 12$. ``10-10-10-10'' means 4 heads are applied and each head predicts the control points with a size of $10 \times 10$. Other settings are also presented in this form. ``Ours'' denotes the combination of ``10-12-14-16'' and residual flow.}
\label{tab:control_points}
%\vspace{-3mm}
\footnotesize
\begin{tabular}{l|cccccc}
\hline
Metrics & {\footnotesize Baseline} & {\footnotesize 12-12-12-12} & {\footnotesize 14-14-14-14} & {\footnotesize 16-16-16-16} & {\footnotesize 10-12-14-16} & Ours\\
\hline
\hline
PSNR & 20.29 & 20.38 & 20.42 &20.02 &20.48 &20.84\\
SSIM & 0.6279 & 0.6311 & 0.6406 &0.6148 &0.6418 &0.6572\\
\hline
\end{tabular}
\end{center}
\footnotesize
%\vspace{-0.4cm}
\end{table}

\noindent \textbf{Motion Estimation}.
It is challenging to estimate multiple motions in one model since the motion's complexities and patterns significantly differ across various tasks. For this purpose, we proposed a flexible and hierarchical architecture to disentangle the motion estimation at the region level and pixel level. As shown in Fig.~\ref{fig:motion_ablation}, better localization performance of the formed mesh can be achieved by increasing the number of TPS points. Besides, the pixel-level residual flow can provide a higher degree of freedom for the motion than only the region-level motion representation, improving the warping results, particularly in the image boundaries and details. Table~\ref{tab:control_points} quantitatively demonstrates the effectiveness of the proposed hierarchical motion estimation module. We also found the upper bound occurs when continuously increasing the number of control points, \textit{e.g}, the performance of four motion estimation heads to predict the size of $16\times16$ TPS is even worse than the size of $12\times12$ TPS. This suggests the performance of multiple-in-one image warping would be limited without proper decoupling of motion manner.

\begin{table}[t]
\setlength{\tabcolsep}{3pt}
\begin{center}
\caption{Ablation study on different task classifiers for the multiple-in-one image warping framework.}
\label{tab:taks_classifier}
%\vspace{-3mm}
\footnotesize
\begin{tabular}{l|cccc}
\hline
Metrics & w/o Classifier& {\footnotesize Classifier-Image} & {\footnotesize Classifier-Point} & Ours\\
\hline
\hline
PSNR & 20.48 & 20.58 & 20.63& 20.83\\
SSIM & 0.6418 & 0.6451 & 0.6463& 0.6558\\
Parameters & - & 3.39M & 0.0592M& 0.0594M\\
\hline
\end{tabular}
\end{center}
\footnotesize
\small
%\vspace{-0.2cm}
\end{table}

\begin{figure*}[t]
\centering
\includegraphics[width=1\linewidth]{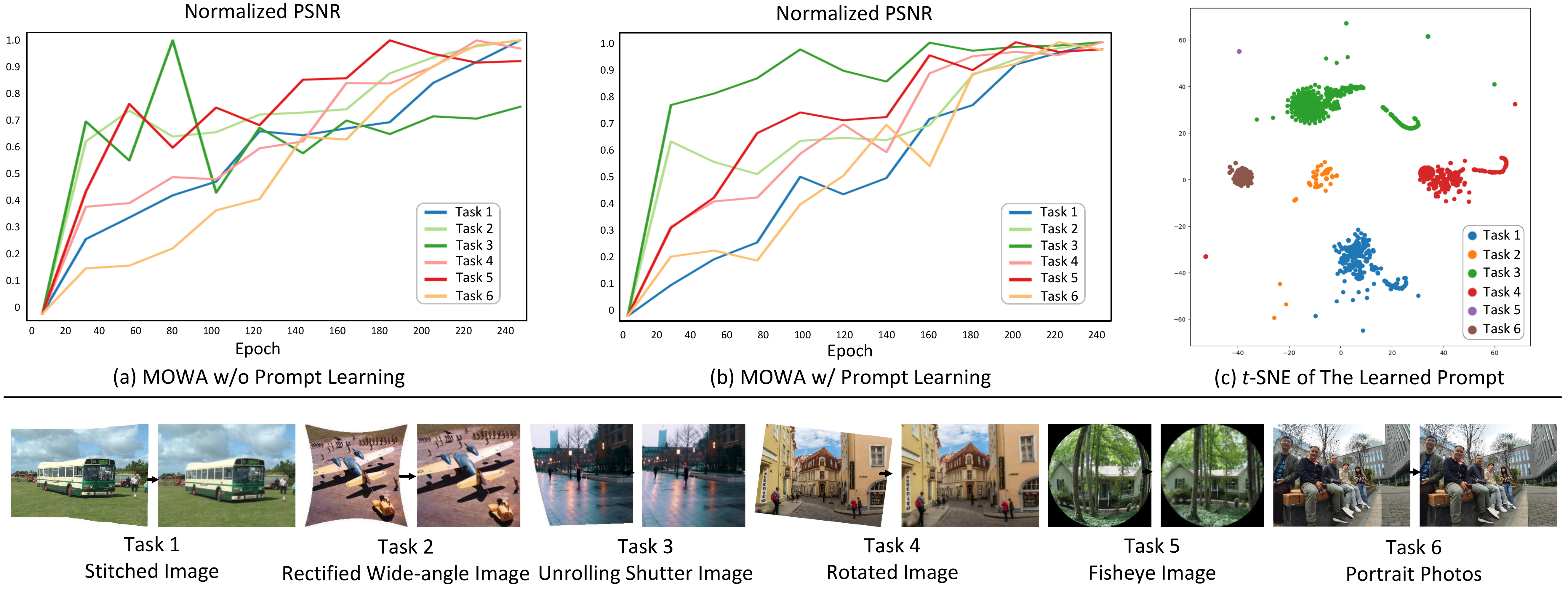}
\caption{Evaluations of the multi-task learning and effectiveness of the proposed prompt learning module. The normalized PSNR and SSIM of validation data are visualized for six tasks. We recap different task types at the bottom.}
\label{fig:prompt_plot}
\end{figure*}

\begin{table}[t]
  \caption{The ablation study of various combinations involved with different warping tasks. We choose the first combination of any two tasks (`Task 1-Stitched Image' with others) as the comparison baseline. The performance gains of PSNR/SSIM are reported for each combination.}
  \setlength{\tabcolsep}{2.4pt}
  \label{tab:combinations}
     \centering
       \footnotesize
   \begin{tabular}{c|c c c c c}
    \hline
      & Task 1 &  Task 2 & Task 3 &  Task 4 &  Task 5 \\
    \hline
    \hline
      Task 1 &- & +0.0/0.0 & +0.27/0.0210 &-0.12/0.0077 &+0.22/0.0198 \\
      Task 2 &+0.0/0.0 & - &+0.32/0.0190  &+0.22/0.0129 &+0.25/0.0151 \\
      Task 3 &+0.0/0.0 & +0.11/0.0077 & - &+0.08/0.034 &+0.19/0.0087 \\
      Task 4 &+0.0/0.0 & +0.26/0.0129 & +0.55/0.0276 &- &+0.47/0.0218 \\
      Task 5 &+0.0/0.0 & +0.98/0.0377  &+0.20/0.0097  &+0.79/0.0285 &- \\
  \hline
\end{tabular}
\end{table}

\vspace{1mm}\noindent \textbf{Task Classifier}.
As the image-based classifier involving redundant image features burdens the main image warping framework, we propose a lightweight point-based classifier to learn the task type from each input image. As listed in Table~\ref{tab:taks_classifier}, four baseline models are designed: image warping model without the task classifier,  with the image-based classifier, with the point-based classifier, and with the point-based classifier compensated with the pooling global features (Ours). Note that we validate different task classifiers only with the TPS prediction modules, showing their direct influence on the region-level motion estimation. The quantitative results demonstrate that we can obtain evident performance gain beyond the vanilla network by adding the task classifier, indicating that the information on task type is meaningful in multi-task training.  

\begin{figure*}[t]
    \centering
    \includegraphics[width=0.9\linewidth]{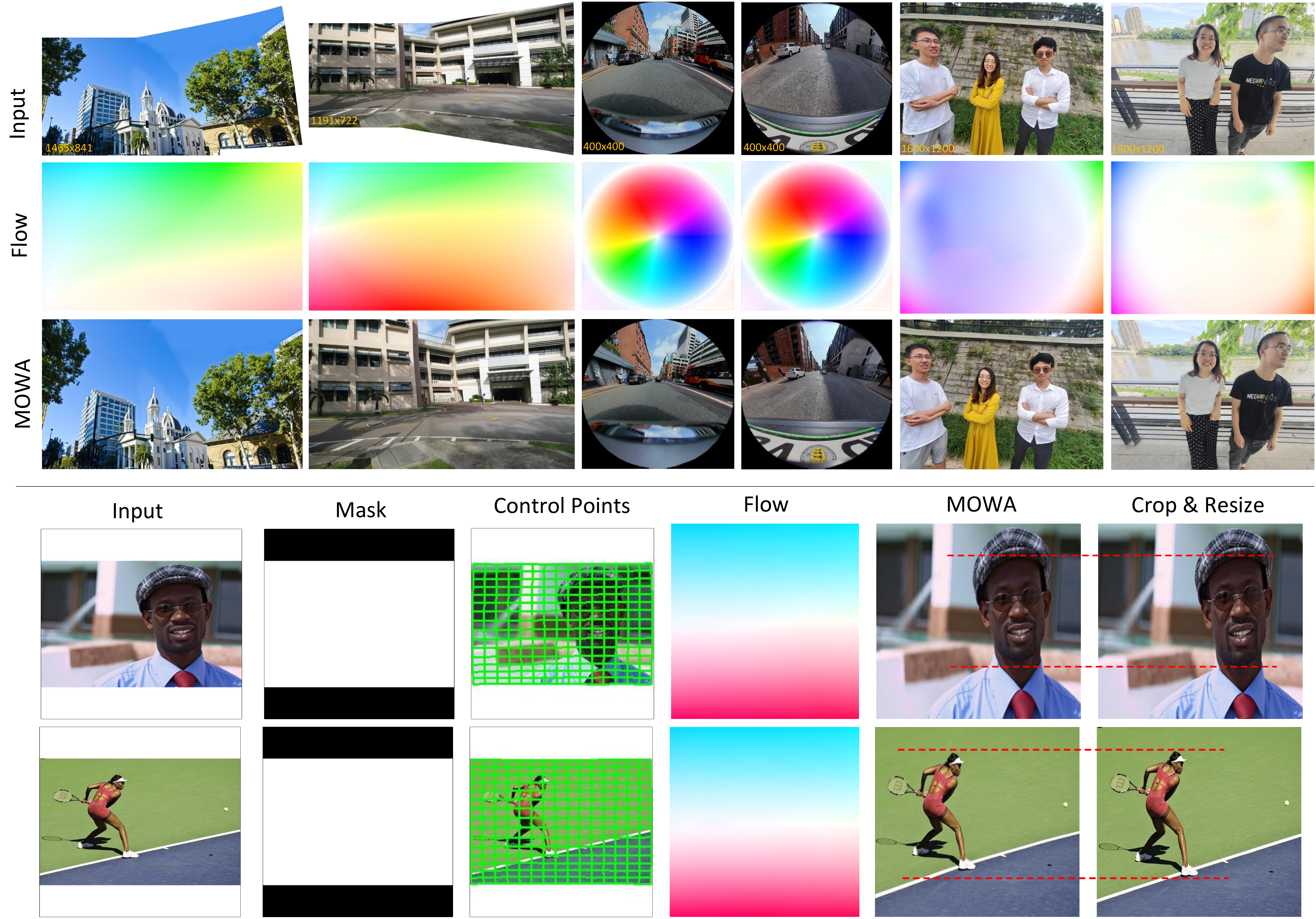}
    %\vspace{-0.2cm}
    \caption{Generalization evaluation of the proposed method: cross-domain evaluation (top) and zero-shot evaluation (bottom). The visualized flow and control points are both predicted by MOWA. The image resolution is marked for each real-world image. In image retargeting results, the red dotted lines measure the stretch extent of the face or body by warping operations.}
    \label{fig:generalization}
    %\vspace{-0.2cm}
\end{figure*}

\subsection{Analysis on Multi-Task Learning and Prompt Learning}
\label{sec:experiment_multitask}
In the proposed method, a prompt learning module is designed to modulate the feature map with the learned task label, which helps to dynamically navigate its extensive parameter space to achieve task-aware image warping. By combining this module into MOWA, the averaged PSNR metric of all warping results gains $+0.21$dB improvements beyond the baseline.

To further analyze the influence of multi-task learning and the effectiveness of the proposed prompt learning, we visualize the quantitative metrics of the warping results of validation images for each task. As illustrated in Figure~\ref{fig:prompt_plot}, the values of normalized PSNR and SSIM are plotted along different training epochs. Considering the ranges of PSNR and SSIM of different warping tasks significantly differ from each other, we normalize all values into the range of $[0, 1]$ to eliminate the data bias. For the portrait photos, we leverage the available warping flow in its training dataset and obtain the corrected images to compute the PSNR and SSIM. For other tasks, we directly use the ground truth of warped images in the corresponding datasets.

\begin{figure}[t]
\centering
\includegraphics[width=1\linewidth]{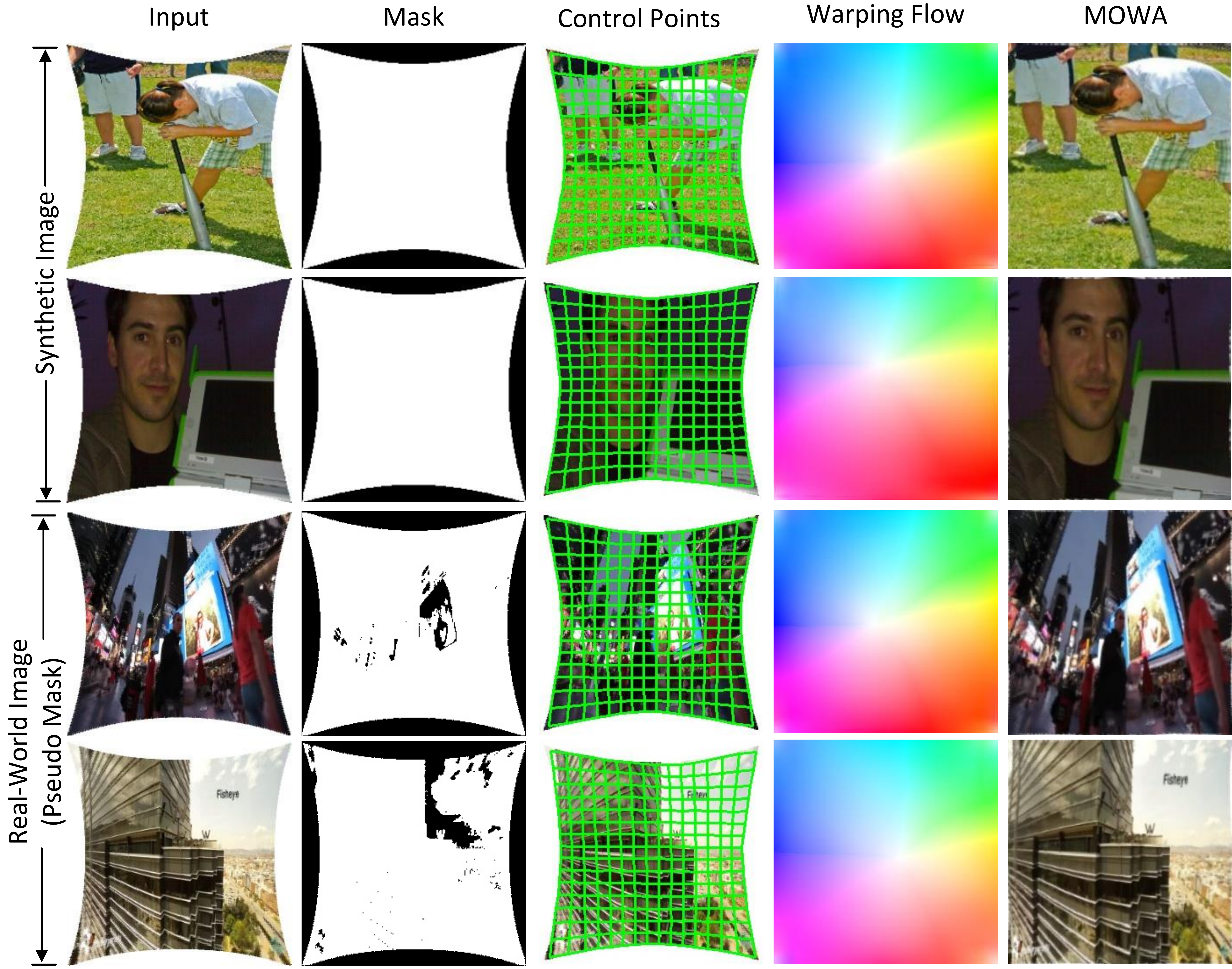}
\caption{While our method is trained on the synthetic datasets with clean masks, it shows good robustness to the real-world data with noisy pseudo masks. The motion estimations of real-world images exhibit structurally reasonable distributions similar to those of the synthetic images.}
\label{fig:real_rec}
\end{figure}

\begin{figure*}[t]
    \centering
    \includegraphics[width=0.98\linewidth]{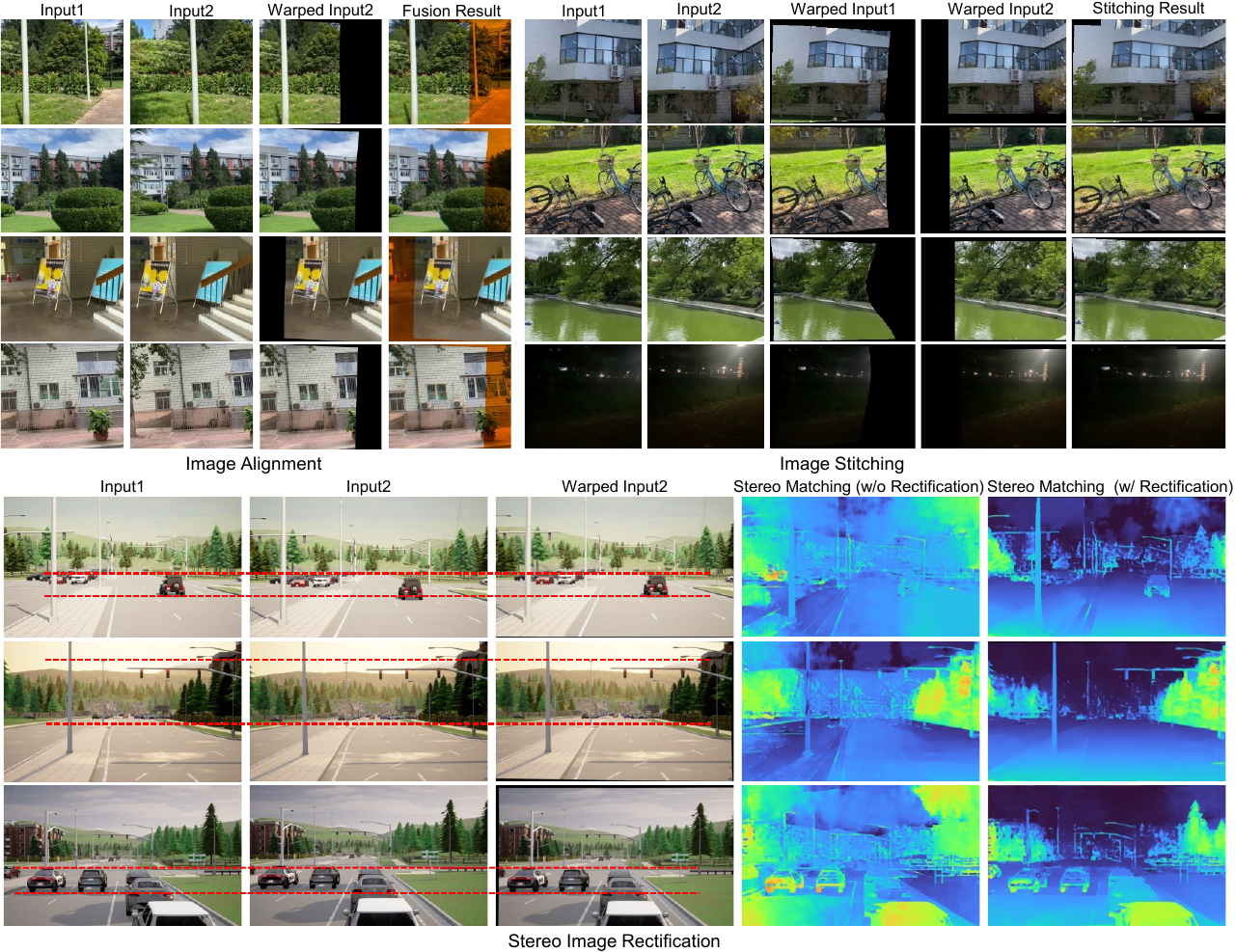}
    %\vspace{-0.2cm}
    \caption{Extended multi-view applications of the proposed method: image alignment, image stitching, and stereo image rectification. In the fusion results of image alignment, the non-overlapping regions of two views are marked in orange. The linearly average fusion algorithm is applied for the image stitching. In the stereo image rectification, the red dotted lines are drawn to highlight the vertical alignment. The stereo matching results are inferred by HITNet~\cite{tankovich2021hitnet}.}
    \label{fig:cross_view_eva}
    %\vspace{-0.2cm}
\end{figure*}

From Figure~\ref{fig:prompt_plot} (a), we have the following three observations: (1) Different tasks show various levels of difficulty when training a multiple-in-one image warping model. For example, learning to warp the unrolling shutter image (task 3) is generally easier than other tasks, which shows the fastest convergence at the first 80 epochs. The reason is that the structures of unrolling shutter images are basically regular, where more than two boundaries are straight and do not need to be warped. On the contrary, learning to warp the stitched images (task 1), rectified wide-angle images (task 2), and portrait photos (task 6) is more challenging (slow convergences can be observed) since their boundaries or expected motions vary greatly in datasets. Especially in the portrait photo dataset, the numbers, shapes, and locations of the faces are quite diverse. (2) The multiple-in-one model tends to sacrifice the performance of some individual tasks to achieve the improvement of holistic warping performance. Particularly, the accuracy of warping the unrolling shutter image is dramatically reduced since the 80th epoch, but the performance of other tasks continues to improve. Such a trade-off among different tasks facilitates an overall improvement but leads to performance conflicts between some tasks. (3) The relationship of different tasks can be positive or negative. For example, the curves of the stitched image and rotated image show consistent trends during the MOWA's training as they share the similar warping principle, \textit{i.e.}, rectangling the irregular image boundaries and keeping the content unchanged. Thus, meaningful interactions between these two tasks could happen if learning a multiple-in-one model. However, the curve of warping the fisheye image shows a converse trend to those of the stitched image and rotated image. The reason for this is that correcting the radial distortion present in fisheye images significantly alters the scene's layout while leaving the image boundaries almost untouched. This approach contrasts with the foundational principles of tasks based on image rectangling. As a result, the imbalanced convergences can be noticed across these tasks with negative relationships. Moreover, the ablation study of exploring various combinations involved with different warping tasks is shown in Table~\ref{tab:combinations}. Specifically, we choose the first combination of any two tasks (`Task 1' with others) as the comparison baseline and report the performance gains of PSNR/SSIM for each subsequent combination.

To dynamically boost the task-aware image warping in a single model, we propose a prompt learning module and a point-based task classifier. As shown in Figure~\ref{fig:prompt_plot} (b), the performance conflicts of different tasks are relieved by prompt learning. All tasks show a similar improvement trend as the training epoch increases, without dramatic performance degradation in certain individual tasks. More importantly, the multiple-in-one framework achieves the unified and best warping performance on all tasks at the end of training epochs. This phenomenon suggests that the framework knows to discriminate and warp different input types using the learned task-specific prior knowledge. Our designed prompts enable MOWA to efficiently traverse and harness its vast parameter space to meet various warping requirements.

We also visualize the \textit{t}-SNE of the learned prompts of MOWA in Figure~\ref{fig:prompt_plot} (c). We can observe these prompts are well-clustered according to the task types. This clear clustering demonstrates the ability of the prompts to learn and represent discriminative motion context, which significantly aids in holistic image warping. The visualization underscores the effectiveness of our approach in capturing and leveraging task-specific features to enhance model performance.

\begin{figure}[t]
\centering
\includegraphics[width=1\linewidth]{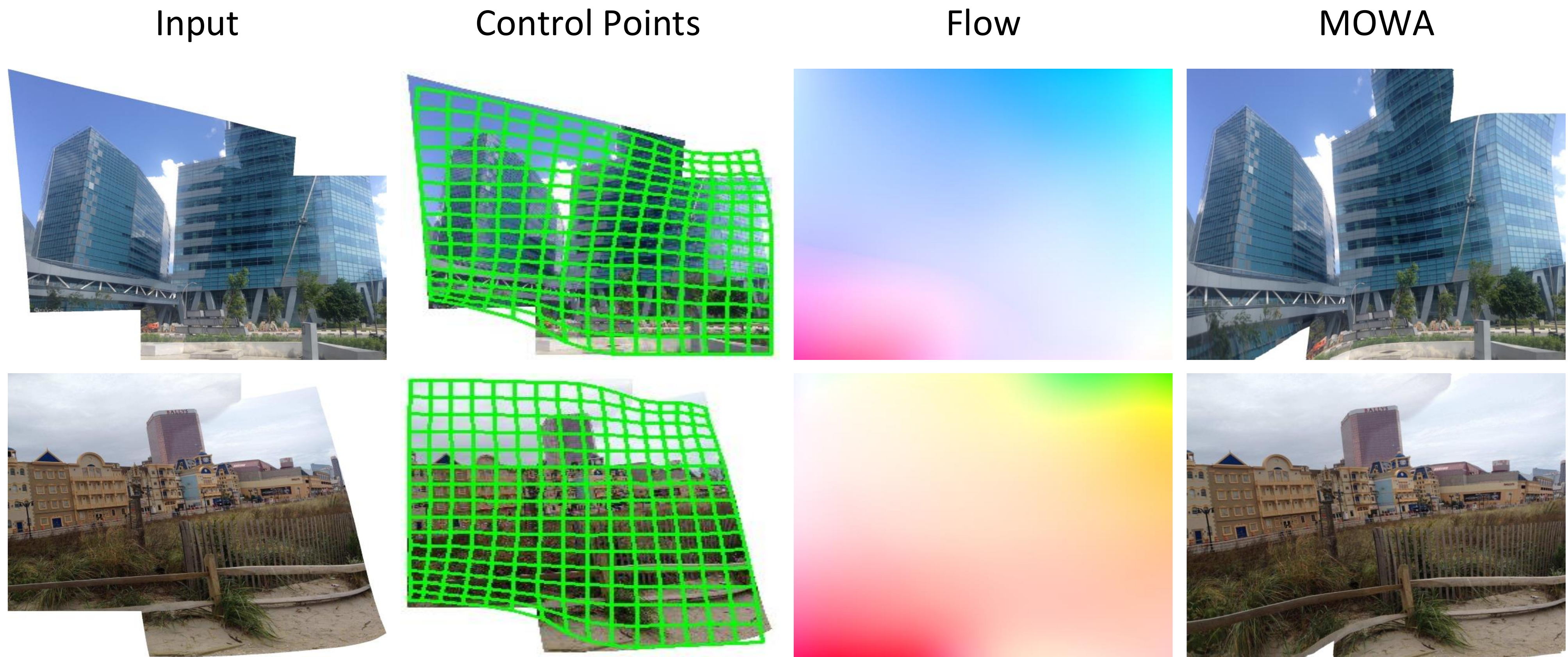}
\caption{Failure cases of the proposed method. It fails to accurately warp the input image with challenging image boundaries using a certain number of control points.}
\label{fig:limitation}
\end{figure}

\subsection{Generalization Evaluation}
We show the generalization ability of the proposed method in terms of the cross-domain and zero-shot evaluations in Fig.~\ref{fig:generalization}. For the cross-domain evaluation, the new inputs belong to the above six practical image warping tasks but they are captured in real-world settings with different cameras, resolutions (from 256$\times$256 to 4K), and scenes. For the zero-shot evaluation, we consider a new image warping task, \textit{i.e.}, image retargeting, which aims to flexibly change the image scale without distorting the content as much as possible. 

The visualization results demonstrate that MOWA can well extend to real-world scenarios, though its training datasets are mostly synthesized by hand-crafted camera models or manipulation spaces. One possible reason is the multiple-in-one model can naturally address the overfitting issue on specific datasets by learning various tasks. In addition to the cross-domain evaluations, we find that while our model does not involve the image retargeting task during training, it is still able to warp the image based on the ``content-aware'' principle. The content-awareness of the proposed warping model is learned from the training datasets in a data-driven manner, in which the labels are constructed by optimizing an energy function with line-preserving mesh deformation or gradient magnitude~\cite{he2013content, he2013rectangling, avidan2023seam}. As we can observe, the predicted control points are accurately aligned to the image boundaries. Such knowledge transferring to new tasks potentially benefits from the shared motion perception across different tasks. Therefore, our results show fewer geometrical distortions for the foreground than the crop and resize operation, with the face and body experiencing less stretching. It is noticed that our method also exhibits satisfactory robustness to noisy data. For instance, in Figure~\ref{fig:real_rec}, although MOWA is trained on the synthetic dataset with clean masks, the warping results of the real-world dataset with noisy pseudo masks (the clean masks are not available in some practical applications) are still structurally reasonable.

\subsection{Extended Applications}
While the proposed method is originally designed for single-image warping tasks, it can be extended to multi-view image warping applications. In particular, we add a feature matching module to the framework, \textit{e.g.}, cost volume, and then fine-tune the pre-trained MOWA on the multi-view warping tasks such as image alignment~\cite{nie2021depth, liu2022content}, image stitching~\cite{nie2021unsupervised}, and stereo image rectification~\cite{kumar2024FGOSrect}. The visual results of MOWA are illustrated in Figure~\ref{fig:cross_view_eva}. As we can observe, our method is capable of addressing image alignment and image stitching problems under different viewpoints, contrasts, and illuminations. Please note that the linearly average fusion algorithm is applied for the stitching. Better fusion results can be achieved by using other advanced composition strategies like seam cutting, but the scope of this work primarily focuses on the performance of image warping. Additionally, the proposed method can rectify the rotations of uncalibrated stereo images under different scenes. The stereo matching results (inferred by HITNet~\cite{tankovich2021hitnet}) of our rectified images also show fewer artifacts and more reasonable distributions than those of uncalibrated inputs.

\subsection{Limitation Discussion}
\label{s-5}
We show some failure cases in Figure~\ref{fig:limitation} and find that the image boundaries are more irregular and the expected displacements of warping are more complicated than most samples. Consequently, it is challenging to approximate the accurate motion structure with a certain number of control points. This limitation could be addressed by adding more control points and cascading more TPS regression heads. Scaling up the resolution of the input image could also potentially improve the warping performance of image boundaries and details.

\section{Conclusion}
\label{sec:conclusion}
We have proposed MOWA in this work, the first multiple-in-one image warping framework in the field of computational photography. It considers six representative and practical tasks in one learning model and uses a unified motion representation to achieve various warping purposes. In particular, to mitigate the difficulty of approximating diverse motions of different tasks, we propose to disentangle the motion estimation at both the region level and pixel level. Then, we enable MOWA's explicit task-aware ability by introducing a lightweight point-based classifier. Compared to the common image-based classifier, it can achieve comparable performance while offering significant parameter reduction. Subsequently, we feed the task label predicted by the task classifier into a prompt learning module and further modulate the feature maps in the decoder, which facilitates a single model to efficiently navigate and leverage its extensive parameter space to meet various warping requirements. Comprehensive experiments demonstrate that MOWA outperforms different SotA methods specifically designed for each single task, with an affordable model size. In the future, we plan to empower MOWA with cross-view and cross-modal abilities, aiming to build a foundation model for universal image warping.

\bibliographystyle{IEEEtran}
\bibliography{main}

\vfill

\end{document}